\documentclass{article}

\PassOptionsToPackage{numbers,sort&compress}{natbib}

\usepackage[main, final]{neurips_2026}

\usepackage[utf8]{inputenc} 
\usepackage[T1]{fontenc}    
\usepackage{hyperref}       
\usepackage{url}            
\usepackage{booktabs}       
\usepackage{amsfonts}       
\usepackage{nicefrac}       
\usepackage{microtype}      
\usepackage{xcolor}         
\usepackage{amsmath}
\usepackage{makecell}
\usepackage{tabularx}
\usepackage{array}
\usepackage{graphicx}
\usepackage{wrapfig}
\usepackage{subcaption}
\usepackage{enumitem}
\usepackage{makecell}
\usepackage{multirow}
\usepackage[table]{xcolor}

\usepackage[table]{xcolor}

\definecolor{mygreen}{rgb}{0.1, 0.5, 0.1}

\newcolumntype{C}[1]{>{\centering\arraybackslash}p{#1}}
\title{HAS: Hierarchical Adaptive Frame Sampling for Video Understanding}

\author{%
  {\normalsize
    \textbf{Yuyao Sun}$^{1}$ \quad
    \textbf{Daiqi Shi}$^{2}$ \quad
    \textbf{Shuang Li}$^{1}$ \quad
    \textbf{Liao Zhang}$^{2}$ \quad
    \textbf{Yang Cai}$^{2}$ \quad
    \textbf{Deqing Wang}$^{1}$
    \thanks{Corresponding author.}
  } \\[2pt]
  {\normalsize
    $^{1}$Beihang University \qquad
    $^{2}$Independent Researcher
  } \\[2pt]
}

\begin{document}

\maketitle
\begin{abstract}
Due to context-length constraints, most MLLMs cannot process full-length videos and therefore rely on sampling a subset of frames as input.
However, existing sampling methods, ranging from uniform sampling to relevance-based selection, are often driven by a single sampling principle and thus struggle to accommodate the heterogeneous evidential requirements of different queries: some demand a holistic understanding of how the video unfolds over time, while others hinge on fine-grained events within a short temporal window.
To address this limitation, we propose \textbf{Hierarchical Adaptive Frame Sampling (HAS)}, a two-stage frame sampling framework. In the first stage, \emph{Backbone Frame Construction}, we apply a Determinantal Point Process (DPP) to sample frames that are both query-relevant and non-redundant. 
These selected frames capture key moments across the video and form the backbone of the sampling set, providing a foundation for subsequent enrichment.
In the second stage, \emph{Adaptive Contextual Enrichment}, we analyze the temporal distribution of the backbone frames to infer the query's requirements and adaptively allocate the remaining frame budget between Local and Global Context. The Local Context enriches the backbone with fine-grained temporal dynamics and short-range causal relations, 
whereas the Global Context connects temporally isolated evidence and provides a holistic view of the entire video to support global understanding.
Through such a hierarchical two-stage process, HAS effectively addresses diverse query requirements. Incorporated into three leading MLLMs, it demonstrates consistently superior performance across the Video-MME, LongVideoBench, and MLVU benchmarks.
\end{abstract}

\section{Introduction}

Multi-modal Large Language Models (MLLMs) have demonstrated impressive performance across a wide range of image perception and reasoning tasks~\cite{blip2,instructblip,llava_1_5,llavanext-interleave,qwen2.5vl,gemini2.5}. 
Extending these capabilities to videos is a natural next step, yet the sheer volume of visual information in videos makes this transition far from trivial. 
Most current video-based MLLMs~\cite{videochatflash,longvu,videoxl,videollama3} follow an image-based paradigm: encoding frames into visual tokens using a vision encoder, and jointly processing these tokens with text tokens in a Large Language Model. 
However, the constrained context window of LLMs precludes processing all video frames simultaneously; typically, a fixed subset (\textit{e.g.}, 32 or 64) is chosen to represent the entire video. 
Thus, the frame sampling strategy becomes a pivotal factor influencing the model's understanding.

A common strategy is uniform sampling, which selects frames at fixed intervals. While this approach preserves coarse global coverage of the storyline, it can easily miss brief yet critical moments. To alleviate this issue, relevance-based sampling, as adopted in SeViLA~\citep{sevila} and MVU~\citep{mvu}, selects query-relevant frames more effectively. However, this focus on relevance often produces visually redundant selections and weakens the narrative context needed for coherent video understanding. 
Taken together, these methods highlight a shared limitation: existing approaches are typically governed by a \emph{single} sampling principle---either emphasizing global semantic (uniform sampling) or prioritizing key evidence (relevance-based sampling). Such a one-sided strategy is ill-suited to the heterogeneous nature of video queries. Some questions demand holistic understanding of the overall narrative (e.g., ``Why does the mother bird bring a fish to the fox?''), whereas others require fine-grained inspection of a few decisive events (e.g., ``How did the cat manage to open the door?''). Since a single principle cannot simultaneously satisfy these competing demands, current methods struggle to adapt across query types, motivating the need for a more adaptive sampling approach.

To address this limitation, we propose \textbf{Hierarchical Adaptive frame Sampling} (HAS), a hierarchical framework that tailors frame selection to the specific requirements of each query.
HAS proceeds in two stages. 
In \textbf{Frame Backbone Construction}, we cast frame sampling as a Determinantal Point Process (DPP). We build the DPP kernel by injecting frame relevance into the inter-frame similarity matrix, and perform greedy MAP inference to select a set of backbone frames that are both highly relevant and non-redundant. This relevance--diversity trade-off encourages the backbone to cover key moments throughout the video, providing a reliable basis for subsequent enrichment.  
In \textbf{Adaptive Contextual Enrichment}, we allocate the remaining budget to complement the backbone with two types of context. \textbf{Local Context} samples temporally neighboring frames around each backbone frame to recover fine-grained dynamics and short-term causal cues, whereas \textbf{Global Context} uniformly samples along the full timeline to maintain a holistic narrative view.

The adaptivity of HAS comes from how it distributes the remaining budget between Local and Global Context according to the temporal layout of the backbone frames. We believe that the temporal layout of the backbone frames provides a useful cue for identifying the query's evidential needs.
When backbone frames form a dense temporal cluster, the queried content is likely concentrated within a short window; HAS therefore emphasizes Local Context to enrich these key moments with nearby evidence. 
Conversely, when backbone frames are sparse and widely spread, the queried subject is more likely to recur across the video and require long-range temporal reasoning; HAS thus assigns more budget to Global Context to connect distant events and maintain narrative coherence. 
In this way, HAS can flexibly shift between zooming in on decisive moments and constructing a holistic overview, addressing the rigidity of one-sided sampling strategies.

To validate HAS, we integrate it into three representative MLLMs: Qwen2.5-VL~\citep{qwen2.5vl}, InternVL3~\citep{internvl3}, and LLaVA-Video~\citep{llavavideo}, and evaluate on three widely-used benchmarks: Video-MME~\citep{videomme}, LongVideoBench~\citep{longvideobench}, and MLVU~\citep{mlvu}. Together, these benchmarks span diverse video lengths and query types, enabling a rigorous evaluation of effectiveness and generalization. Experimental results show that HAS not only significantly outperforms the default uniform sampling strategy used by these models but also surpasses other advanced sampling methods, such as AKS~\citep{AKS} and TSTAR~\citep{tstar}.

\vspace{-1ex}

\section{Related Work}
\subsection{Multimodal Large Language models}

The remarkable success of Large Language Models (LLMs) has motivated extending them to incorporate visual information, giving rise to Multimodal LLMs (MLLMs). Early models relied on heavyweight vision--language alignment designs. For instance, BLIP-2~\citep{blip2} and Qwen-VL~\citep{qwenvl} adopt a Q-Former to bridge visual embeddings into the language embedding space, while Flamingo~\citep{flamingo} injects cross-attention layers into the LLM to fuse visual and textual representations. LLaVA~\citep{llava} later shows that effective vision--language alignment can be achieved with a much simpler design, using only a lightweight MLP projector and avoiding substantial architectural changes. This paradigm has since been widely adopted by other MLLMs, enabling strong visual understanding performance in models such as MiniGPT-4V~\citep{minigpt4}, ShareGPT4V~\citep{sharegpt4v}, Qwen2.5-VL~\citep{qwen2vl} and InternVL3~\citep{internvl3}.

\subsection{Video-based MLLMs}
Extending MLLMs from images to videos introduces a key challenge: videos contain far more information, and encoding all frames into tokens would quickly exceed the LLM's context window. Most video-based MLLMs therefore build upon image-based architectures and sample a subset of frames as input, as in Video-ChatGPT~\citep{video-chatgpt}, Video-LLaMA~\citep{video-llama}, Deepseek-VL2~\citep{deepseek-vl2}, and Video-LLaVA~\citep{video-llava}. Inevitably, this discards information from unsampled frames, and the issue becomes more severe as video length increases. 
To mitigate this, methods such as ADACM~\citep{adacm} and HierarQ~\citep{hierarq} employ transformer-based designs to process frames sequentially and compress the entire video into a compact representation. While this reduces the information loss induced by sampling, such architectures generally underperform mainstream sampling-based MLLMs.

\subsection{Frame sampling for video MLLMs }

Given the limited context window of MLLMs, frame sampling has emerged as the dominant solution for adapting long videos to VideoQA tasks, making it a critical factor in model performance. Early influential methods such as SeViLA~\citep{sevila} and MVU~\citep{mvu} focused on query relevance by using MLLMs to assign importance scores to each frame individually. To address the efficiency bottleneck of scoring each frame individually, \citet{MLLMBASED} predict all frame-level scores in a single forward pass. Other concurrent approaches deviate from explicit query-relevance scoring at inference: Koala~\citep{koala} employs fixed sparse sampling to condition subsequent visual processing, while ViLA~\citep{vila} trains an end-to-end module to learn a sampling strategy. More recent methods refine the selection criteria: AKS~\citep{AKS} simultaneously considers both the relevance of a frame and the temporal coverage of the already selected frames, and TSTAR~\citep{tstar} converts temporal search into spatial search by arranging frames into a large grid image. 
However, existing methods often suffer from a key limitation: they are typically governed by a single sampling principle (e.g., uniform coverage or relevance-only selection) and thus struggle to adapt to the heterogeneous evidential requirements of different query types. To the best of our knowledge, we are the first to explicitly highlight this query-dependent demand and propose \textbf{Hierarchical Adaptive frame Sampling} (HAS) to address this challenge.

\vspace{-1ex}

\section{Method}
\subsection{Preliminaries: Problem Formulation}

To motivate our method, we first formalize the video question answering pipeline. A video $V=\{v_1,\dots,v_T\}$ is a sequence of $T$ frames. Given the limited context window of an MLLM $\mathcal{M}$, it is typically infeasible to encode all $T$ frames; instead, the model operates on a subset of $K$ frames with $K \ll T$. We denote the sampling strategy as $\mathcal{S}$, which takes the video $V$ and optionally the query $Q$ as input and returns a set of $K$ frame indices, $\mathcal{I}=\mathcal{S}(V,Q)$. The sampled frames $\{v_i\}_{i\in\mathcal{I}}$, together with $Q$, are then fed into $\mathcal{M}$ to generate an answer $A$. Consequently, the answer quality depends critically on the effectiveness of $\mathcal{S}$. Ideally, an optimal strategy $\mathcal{S}^*$ would select indices $\mathcal{I}^*$ that maximize the probability of generating the ground-truth answer $A_{\text{gt}}$:
\begin{equation}
    \mathcal{I}^* = \arg\max_{\mathcal{I} \subset \{1..T\}, |\mathcal{I}|=K} P(A=A_{\text{gt}} \mid \{v_i\}_{i \in \mathcal{I}}, Q; \mathcal{M}).
    \label{eq:ideal_objective}
\end{equation}
However, learning such a strategy is intractable, as even finding $\mathcal{I}^*$ for a single instance would require exhaustively evaluating all possible subsets with the MLLM. The goal of this work is therefore to design a practical and effective strategy, $\mathcal{S}_{\text{ours}}$, that serves as a proxy to $\mathcal{S}^*$.

\subsection{Hierarchical Adaptive frame Sampling}

\begin{figure*}[t]
    \centering 
    \includegraphics[width=1\textwidth]{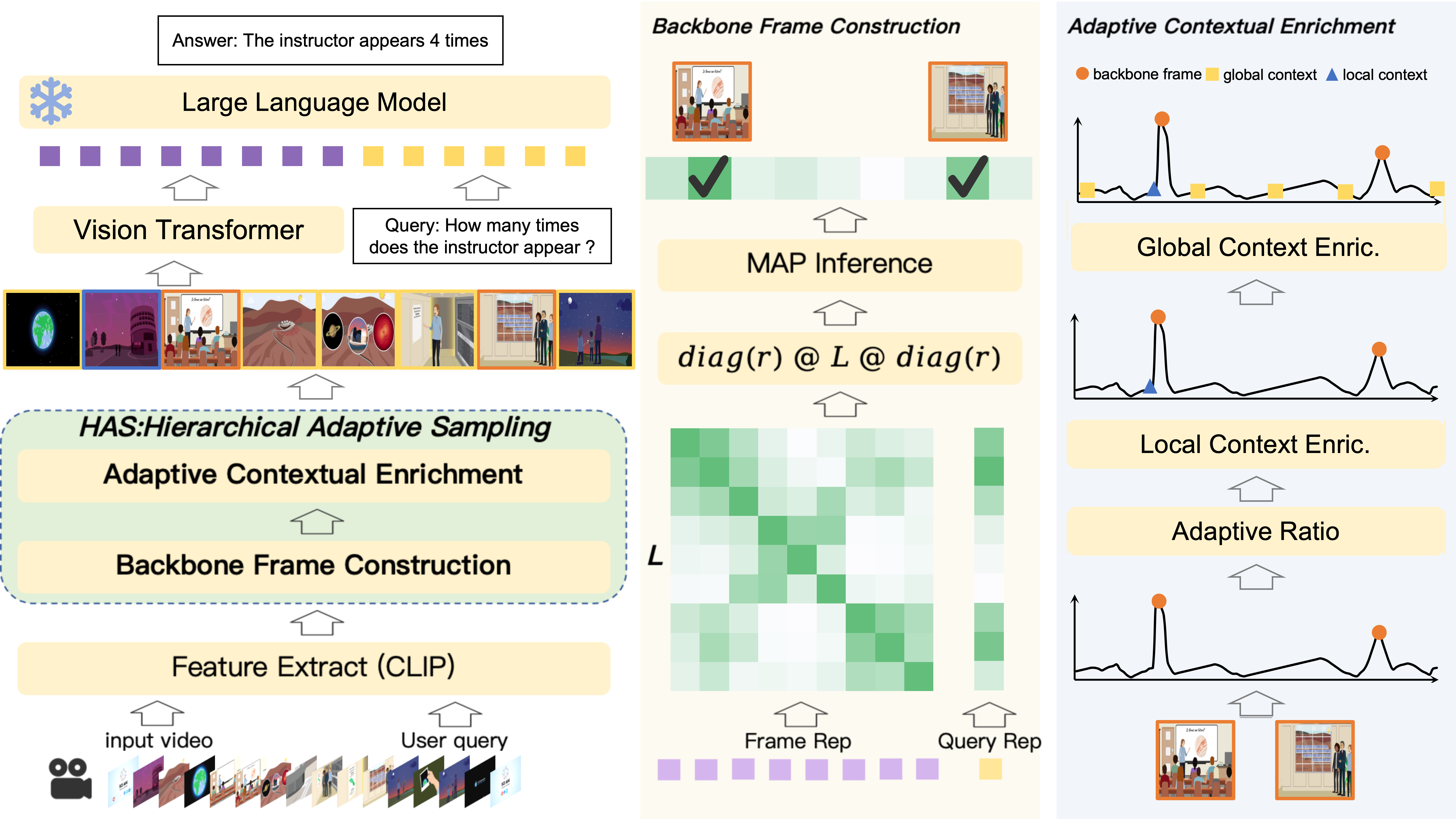} 
    \caption{The overall flowchart of HAS. HAS is a training-free approach that enhances model performance solely by modifying the frames sampled from the video. In the Adaptive Contextual Enrichment module, different colors denote different frame types: orange for Backbone Frames, green for Global Context, and blue for Local Context.} 
    \label{fig:my_figure} 
    \vspace{-1em}
\end{figure*}


Motivated by the need to tailor evidence to different VideoQA queries, we introduce HAS, a two-stage frame sampling framework. 
In \textbf{Backbone Frame Construction}, HAS selects query-relevant yet diverse frames to form an informative backbone. 
In \textbf{Adaptive Contextual Enrichment}, it adaptively allocates the remaining budget between Local and Global Context to complement the backbone based on the needs of the query.
The following sections detail each stage.

\subsubsection{Backbone Frame Construction}
\label{sec:dpp_selection}
HAS begins by selecting a set of backbone frames that are simultaneously highly relevant to the query and internally diverse. We model this as a subset selection problem using a Determinantal Point Process, a natural choice for balancing quality and diversity. The effectiveness of DPP has been well established in tasks like recommendation systems~\citep{fastdpp,dpprec} and, more recently, token pruning~\citep{dpp}.

In our formulation, we construct a kernel matrix $L$ that jointly incorporates frame-query relevance (quality) and inter-frame redundancy (diversity) to guide backbone frame selection.
Specifically, we use a pre-trained CLIP model to extract embeddings for each video frame $\mathbf{f}_i$ and the query, and compute a relevance score $r_i$ via cosine similarity. 
The kernel entries are then defined as:
\begin{equation}
    L_{ij} = r_i \cdot \text{sim}(\mathbf{f}_i, \mathbf{f}_j) \cdot r_j,
\end{equation}
where $\text{sim}(\cdot, \cdot)$ denotes the cosine similarity between frame embeddings.

Given $L$, DPP assigns each candidate subset $\mathcal{I}$ a probability proportional to the determinant of its corresponding submatrix, $\det(L_\mathcal{I})$, which simultaneously reflects the collective relevance of the selected frames and their diversity. 
Selecting the backbone frames can thus be formulated as a maximum a posteriori (MAP) inference problem under the DPP model, i.e., finding the subset of size $k_b$ that maximizes:
\begin{equation}
    \mathcal{I}_{\text{backbone}} = \underset{\mathcal{I} \subseteq \{1..T\}, |\mathcal{I}|=k_b}{\arg\max} \det(L_\mathcal{I}),
    \label{eq:map_inference}
\end{equation}
where $k_b$ denotes the number of backbone frames to be selected.

Since this optimization is NP-hard, we employ an efficient greedy MAP inference algorithm from~\citet{fastdpp}. 
This algorithm iteratively selects the sample that provides the largest marginal gain to the log-determinant until the subset size is reached. 
To prevent a fixed budget from forcing the selection of low-relevance frames, we introduce an adaptive stopping criterion: during this process, we terminate when the marginal gains of adding any remaining candidate frame all fall below a predefined threshold~$\eta$. We denote the resulting backbone frame indices as $\mathcal{I}_{\text{backbone}}$, which are then used in the subsequent stage.


\subsubsection{Adaptive Contextual Enrichment}
\label{sec:context_enrichment}

The backbone frames provide a foundation basis that captures the key moments across the entire video, but remains incomplete on its own due to limited surrounding context.
To enrich this foundation for more complex reasoning, we utilize the remaining frame budget, 
$k_{\text{context}} = K - |\mathcal{I}_{\text{backbone}}|$, to sample additional context frames. This budget is adaptively allocated between two types: Global Context and Local Context.

\paragraph{Local Context.} 

While the backbone frames pinpoint critical moments, the DPP-enforced diversity often yields temporally sparse snapshots that lack the immediate context needed to recover local dynamics and short-term causal cues. 
To remedy this, we apply a ripple expansion process to sample Local Context frames, denoted as $\mathcal{I}_{\text{LC}}$. 
We first sort the backbone frames by relevance (descending) and push them into a FIFO queue, treating each queued frame as an expansion point. 
We then repeatedly pop an expansion point and scan neighboring frames outward along the temporal axis in both directions .
The search stops upon encountering either an already selected frame or an unselected frame whose similarity to the expansion point falls below $\tau$. 
All newly qualified frames are added to $\mathcal{I}_{\text{LC}}$ and also enqueued as new expansion points, allowing the expansion to propagate to nearby regions. 
We stop this stage once $\lvert \mathcal{I}_{\text{LC}} \rvert$ reaches the Local Context budget $k_{\text{LC}}$. 
Overall, this procedure augments the backbone with temporally adjacent evidence, capturing local dynamics and short-range causal relations around key moments.


\paragraph{Global Context.} 

Some queries require a coherent understanding of how the video unfolds over time, which backbone frames alone may not fully capture.
To address this, HAS allocates part of the frame budget to \emph{Global Context} and samples an additional set of frames $\mathcal{I}_{\text{GC}}$ to provide broad coverage of the entire video. 
We explored multiple Global Context sampling strategies and found that simple uniform sampling performs best.
Accordingly, we adopt uniform sampling as the default Global Context strategy in this work, and provide detailed comparisons and analysis in Section~\ref{ablation_study_GC}.

\paragraph{Adaptive Budget Allocation.}
\label{sec:adaptive_ratio}

A fixed allocation between Global and Local Context is suboptimal, as the appropriate balance is inherently query-dependent. We leverage the temporal layout of the backbone frames as a cue for this allocation. Intuitively, when backbone frames form a dense temporal cluster, the query is likely centered on a short, localized segment, and thus benefits more from Local Context to recover fine-grained dynamics and short-term causal cues. In contrast, when backbone frames are sparse and widely spread, the queried content is more likely to recur throughout the video, calling for broader Global Context to connect distant events and maintain narrative coherence.

To quantify this temporal layout, we compute an entropy-based globalness score $\mathcal{S}_{\text{glob}}$. We partition the video timeline into $B$ equal temporal bins, where we set $B = |\mathcal{I}_{\text{backbone}}|$. Let $s_t$ denote the relevance score of a backbone frame $t \in \mathcal{I}_{\text{backbone}}$. For each bin $b$, we accumulate the weighted mass
\begin{equation}
a_b=\sum_{t\in \mathcal{I}_{\text{backbone}}} s_t\,\mathbf{1}\!\left(t \in b\right),
\end{equation}
where $\mathbf{1}(\cdot)$ equals $1$ if frame $t$ falls into bin $b$ and $0$ otherwise. We normalize $\{a_b\}$ into a discrete distribution $\{p_b\}_{b=1}^{B}$ by
\begin{equation}
p_b=\frac{a_b}{\sum_{j=1}^{B} a_j},
\end{equation}
and compute the normalized entropy
\begin{equation}
\mathcal{S}_{\text{glob}}=\frac{-\sum_{b=1}^{B} p_b \log\left(p_b+\epsilon\right)}{\log B}.
\end{equation}
Finally, we allocate the context budget $k_{\text{context}}$ by assigning $\lfloor \mathcal{S}_{\text{glob}} \cdot k_{\text{context}} \rfloor$ frames to Global Context and using the remaining frames for Local Context. The final set of frame indices passed to the MLLM is
$\mathcal{I}_{\text{final}} = \mathcal{I}_{\text{backbone}} \cup \mathcal{I}_{\text{GC}} \cup \mathcal{I}_{\text{LC}}$.

\vspace{-1ex}

\section{Experiments}
\subsection{Experimental Setup}

\begin{table*}[t!]
  \centering
  \footnotesize
  \setlength{\tabcolsep}{2.8pt}
  \renewcommand{\arraystretch}{0.88}
  \setlength{\aboverulesep}{0.18ex}
  \setlength{\belowrulesep}{0.18ex}
  \setlength{\cmidrulesep}{0.10ex}
  \caption{Main results on long video understanding benchmarks. 
    The LVB-60s subset is excluded from our evaluation since our approach re-samples from an initial 1fps stream.
    \textbf{Bold} indicates the best performance and \underline{underline} indicates the second-best within each group.}
  \label{tab:main_result}
  
  \resizebox{\textwidth}{!}{%
    \begin{tabular}{p{2.5cm} c c C{0.8cm} C{0.8cm} C{0.8cm} C{0.8cm} C{0.8cm} C{0.8cm} C{0.8cm} C{0.8cm}}
      \toprule 
      \multirow{2}{*}{Model} & \multirow{2}{*}{Size} & \multirow{2}{*}{Frames} 
      & \multicolumn{2}{c}{LVB} 
      & \multicolumn{3}{c}{VideoMME} 
      & \multicolumn{3}{c}{MLVU} \\
      \cmidrule(lr){4-5} \cmidrule(lr){6-8} \cmidrule(lr){9-11} 
      & & & 600 & 3600 & Short & Med. & Long & Hol. & S-D & M-D \\
      \midrule 
      \multicolumn{11}{l}{\textit{Proprietary}} \\
      GPT4o     & -- & -- & 66.7 & 61.6 & 77.7 & 66.2 & 59.8 & 79.4 & 53.7 & 35.7 \\
      Gemini-1.5-Flash & -- & -- & 63.1 & 57.3 & 71.3 & 58.6 & 52.1 & 73.9 & 52.5 & 27.9 \\
      \midrule 
      \multicolumn{11}{l}{\textit{Open-source}} \\
      VideoChat2       & 7B  & 16 & 39.0 & 37.5 & 54.2 & 42.8 & 39.9 & 71.7 & 50.0 & 24.9 \\
      Long-VITA        & 14B & 64 & 52.7 & 46.8 & 74.6 & 61.2 & 55.0 & 81.0 & 69.8 & 50.1 \\
      LLAVA-OV         & 72B & 64 & 61.6 & 56.5 & 77.2 & 63.2 & 58.2 & 83.1 & 74.9 & 48.6 \\
      LLAVA-Video      & 72B & 64 & 63.9 & 59.3 & 80.8 & 68.6 & 62.1 & 81.2 & 78.7 & 55.7 \\
      \midrule

      Qwen2.5VL     & 7B & 64 & 59.0 & 51.1 & 73.9 & 62.3 & 52.2 & 80.6 & 66.0 & 43.9 \\
      \quad + TOP   & 7B & 64 & \underline{62.9} & 56.0 & 74.2 & 62.3 & 53.8 & 76.5 & 72.2 & 54.4 \\
      \quad + TStar & 7B & 64 & 62.6 & 55.8 & 74.2 & 63.4 & 50.3 & 78.8 & \underline{72.4} & \underline{55.5} \\
      \quad + AKS   & 7B & 64 & 62.1 & \underline{56.2} & \underline{74.4} & \underline{65.1} & \underline{54.9} & \textbf{82.3} & 71.4 & 51.0 \\
      \quad + HAS   & 7B & 64 & \textbf{63.8} & \textbf{56.4} & \textbf{74.8} & \textbf{65.3} & \textbf{55.4} & \underline{80.8} & \textbf{73.4} & \textbf{57.2} \\
      \midrule

      LLAVA-Video   & 7B & 64 & 59.2 & 50.9 & 75.9 & 63.2 & \underline{52.1} & \underline{79.7} & 75.0 & 53.7 \\
      \quad + TOP   & 7B & 64 & \underline{62.1} & 53.7 & 76.9 & 59.8 & 51.9 & 75.6 & 74.4 & \underline{57.7} \\
      \quad + TStar & 7B & 64 & 61.9 & 53.9 & 76.6 & 63.7 & 51.4 & 76.7 & \underline{75.1} & 56.8 \\
      \quad + AKS   & 7B & 64 & 61.4 & \underline{54.3} & \underline{77.0} & \textbf{65.1} & \textbf{55.0} & 79.3 & 74.9 & 52.7 \\
      \quad + HAS   & 7B & 64 & \textbf{63.1} & \textbf{56.2} & \textbf{77.6} & \underline{64.3} & \textbf{55.0} & \textbf{80.6} & \textbf{76.1} & \textbf{60.6} \\
      \midrule

      InternVL3      & 8B & 64 & 61.7 & 53.0 & 76.2 & 66.4 & \underline{55.4} & \textbf{84.7} & 73.9 & 52.5 \\
      \quad + TOP   & 8B & 64 & 63.6 & \underline{56.2} & \underline{77.8} & 63.6 & 54.6 & 80.8 & 75.4 & 57.0 \\
      \quad + TStar & 8B & 64 & \underline{64.1} & 55.7 & 77.6 & 66.9 & 54.1 & 81.7 & \underline{76.3} & \underline{57.5} \\
      \quad + AKS   & 8B & 64 & 63.3 & 54.8 & 77.2 & \underline{68.9} & 55.3 & \underline{82.3} & 75.0 & 52.9 \\
      \quad + HAS   & 8B & 64 & \textbf{64.6} & \textbf{56.4} & \textbf{78.1} & \textbf{70.0} & \textbf{56.3} & \underline{82.3} & \textbf{77.6} & \textbf{58.1} \\
      \bottomrule 
    \end{tabular}%
  }
\end{table*}

\noindent\textbf{Datasets and Models.}
To empirically validate the effectiveness of HAS, we conduct comprehensive experiments on three widely used video understanding benchmarks: VideoMME~\citep{videomme}, LongVideoBench~\citep{longvideobench}, and MLVU~\citep{mlvu}. These benchmarks are characterized by their diverse range of question types and video lengths, with some videos extending beyond one hour. This makes them a rigorous testbed for evaluating the efficacy of frame sampling strategies. 
For models, we select three powerful MLLMs: Qwen2.5-VL-7B~\citep{qwen2.5vl}, InternVL3-8B~\citep{internvl3}, and LLaVA-Video-7B~\citep{llavavideo}. 

\noindent\textbf{Compared Methods.}
We compare our proposed method, HAS, against four frame sampling methods. These are: 
(1) UNI, which samples frames at fixed temporal intervals across the video's timeline and serves as the default method for the models evaluated in this paper; 
(2) TOP, which selects frames in descending order of their query-relevance scores; 
(3) AKS~\citep{AKS}, an enhancement of TOP that simultaneously considers both the relevance of a frame and the temporal coverage of the already selected set of frames and (4) TSTAR~\citep{tstar}, which select question-relevant frames via object-centric visual search rather than semantic similarity, iteratively narrowing down the temporal region of interest through coarse-to-fine resampling.

\noindent\textbf{Implementation Details.} 
We conduct all our experiments using the VLMEvalKit framework~\citep{vlmevalkit}. HAS is implemented in a plug-and-play manner: we only change which frames are selected as input while keeping the model weights frozen throughout all evaluations. To improve computational efficiency, we first pre-sample a candidate pool from each video at 1~FPS, and then apply our method to select the final set of frames from this pool. For the DPP backbone selection, we construct the candidate set by taking the top-$6K$ frames ranked by relevance, where $K$ is the frame budget. All relevance scores are computed using CLIP-B~\citep{clip}. Unless otherwise specified, we set $\eta=0.03$.

\subsection{Main results}

\subsubsection{Quantitative results}

As presented in Table~\ref{tab:main_result}, our proposed HAS is comprehensively evaluated across three models and three benchmarks. In the great majority of evaluations, HAS outperforms the other sampling strategies, achieving the best performance.

The results on the LVB and VideoMME benchmarks particularly highlight our method's advantage in long-duration video understanding. For instance, on the challenging LVB-3600 benchmark, HAS boosts the LLAVA-Video-7B model's score to 56.2, a clear +1.9 improvement over the next-best method, AKS. Similarly, on the VideoMME Long benchmark, HAS (on InternVL-3) achieves a score of 56.3, significantly outperforming AKS (55.3). Moreover, this strong performance across both LVB, which emphasizes single-scene analysis, and VideoMME, which requires a holistic understanding of the video timeline to answer, showcases our method's ability to adapt to diverse problem types. 

The results on the MLVU benchmark further showcase our method's adaptability. HAS achieves top scores on detail-oriented tasks like Single-Detail (S-D) and Multi-Detail (M-D) (e.g., 77.6 and 58.1 with InternVL3-8B), while also securing highly competitive results on Holistic (Hol.) tasks, including scores of 80.6 with LLAVA-Video and 82.3 with InternVL3, the best and second-best results among all sampling strategies. This demonstrates HAS's dual capability: providing a global perspective for holistic queries, while also capturing the local dynamics and short-term context for questions focused on specific moments.

Notably, HAS enables smaller open-source models to achieve performance competitive with, and in some cases even superior to, much larger or proprietary counterparts. For example, HAS-enhanced LLaVA-Video-7B achieves 60.6 on MLVU M-D, outperforming its 10$\times$ larger 72B counterpart (55.7), demonstrating immense gains at a fraction of the computational cost. In another compelling case, InternVL3-8B with HAS (70.0) outperforms the proprietary GPT-4o (66.2) on VideoMME Med. These results highlight a key insight: in certain scenarios, employing our economical HAS as a video sampling method is a more effective strategy for performance enhancement than the costly alternative of training ever-larger models.

\subsubsection{Qualitative results}
\begin{figure*}[t]
    \centering 
    \includegraphics[width=1\textwidth]{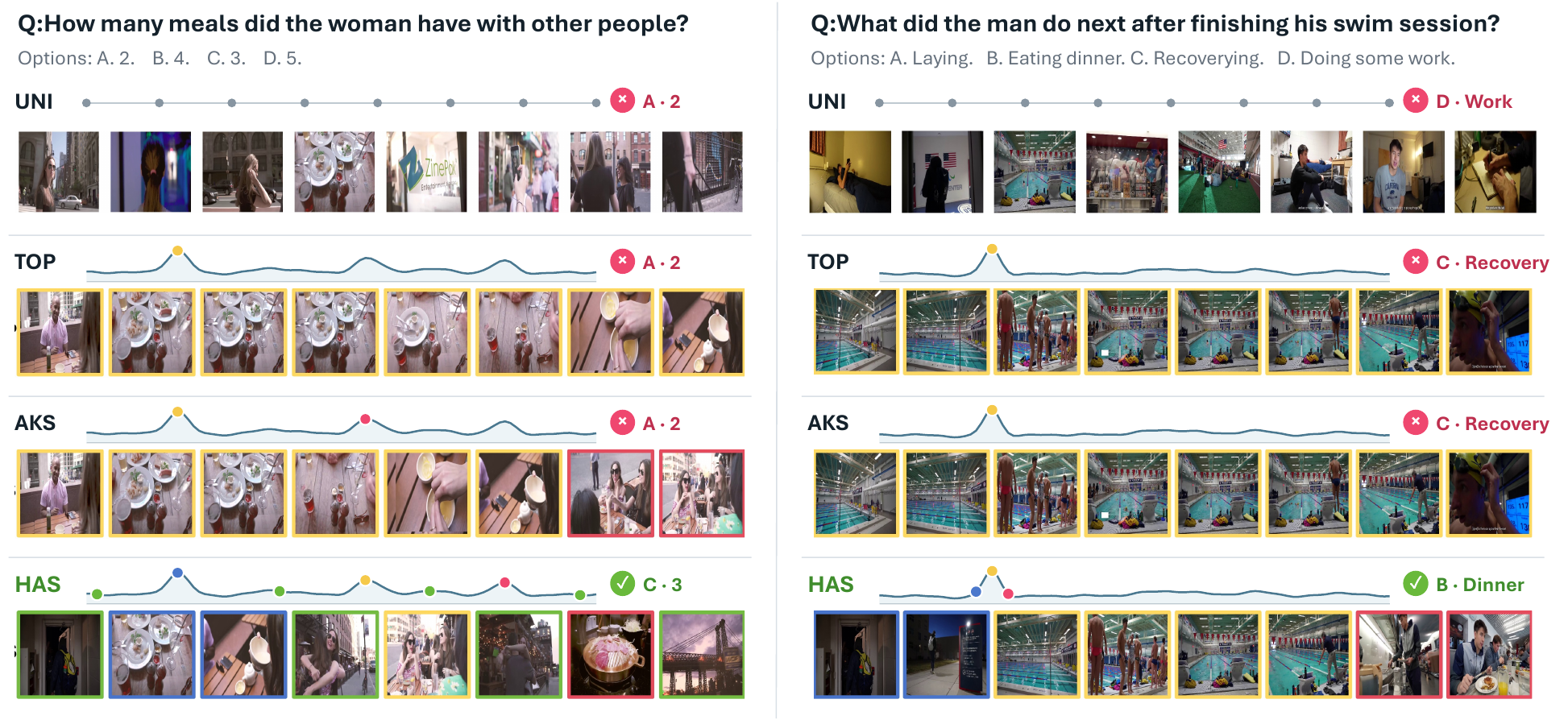} 
    \caption{Two examples of how different sampling strategies impact video understanding. The curve illustrates the video's temporal relevance profile. A frame's border color matches the color of its sampling dot on the curve.} 
    \label{fig:qualitative_comparison} 
\end{figure*}


Figure~\ref{fig:qualitative_comparison} illustrates the behavior of different sampling strategies on two representative cases. In the first case, HAS, benefiting from DPP, accurately captures three distinct question-relevant scenes that are all necessary for answering the question, whereas AKS and TOP ignore some of these key scenes to varying degrees. In addition, the global context provide a more complete view of how the video evolves over time, so that the final sampled frame set offers a coherent and sufficient basis for answering the question. In the second case, HAS enriches the selected query-relevant backbone frames with local context which supplies the crucial short term context required to answer the question correctly. In comparison, AKS and TOP rely solely on key frames, which results in incomplete contextual understanding and incorrect predictions. Jointly, these two cases collectively demonstrate the effectiveness of HAS. We provide additional qualitative examples in the appendix, covering diverse scenarios under different frame budgets.

\subsection{Ablation study}

\subsubsection{Ablation Study On Frame Budgets}

\begin{table*}[t]
  \centering
  \small
  \caption{Ablation study on frame budget for HAS on InternVL3.}
  \label{tab:ablation_study_frames}
  \begin{tabular*}{\textwidth}{@{\extracolsep{\fill}} l c c c c c c c c c}
    \toprule
    Metric & \multicolumn{3}{c}{LVB} & \multicolumn{3}{c}{VideoMME} & \multicolumn{3}{c}{MLVU} \\
    \cmidrule(lr){2-4} \cmidrule(lr){5-7} \cmidrule(lr){8-10}
    Frames & 64 & 32 & 16 & 64 & 32 & 16 & 64 & 32 & 16 \\
    \midrule
    UNI & 59.8 & 56.4 & 55.1 & 66.0 & 64.4 & 61.9 & 73.1 & 70.0 & 65.8 \\
    TOP & \underline{61.8} & \underline{60.9} & 60.0 & 65.3 & 63.5 & 61.0 & \underline{74.6} & 73.0 & \underline{69.4} \\
    AKS & 61.1 & 59.5 & \underline{60.1} & \underline{67.1} & \underline{66.2} & \underline{65.8} & 74.3 & \underline{73.1} & 65.0 \\
    HAS & \textbf{62.2} & \textbf{61.5} & \textbf{60.7} & \textbf{68.1} & \textbf{66.7} & \textbf{66.2} & \textbf{75.8} & \textbf{74.1} & \textbf{71.4} \\
    \bottomrule
  \end{tabular*}
\end{table*}


Table~\ref{tab:ablation_study_frames} presents a performance comparison under varying input frame budgets (64, 32, and 16). The results demonstrate the strong performance and robustness of our method across different frame budgets. Under ample budgets of 64 and 32 frames, HAS consistently achieves the best performance across all benchmarks. For instance, at 32 frames, HAS secures top scores of 66.7 on VideoMME and 61.5 on LVB, clearly surpassing the next-best scores of 66.2 (AKS) and 60.9 (TOP), respectively. When the frame budget is constrained to 16 frames, splitting the budget into three components (DPP backbone, Local Context, and Global Context) may allocate too few frames to each part, limiting their effectiveness under such extreme constraints. To address this issue, we simplify HAS by retaining only the first-stage DPP backbone construction at 16 frames, avoiding over-fragmentation of the limited budget. With this adaptation, HAS remains the best-performing method under the 16-frame setting across all benchmarks, achieving 66.2 on VideoMME, 60.7 on LVB, and 71.4 on MLVU.


\subsubsection{Ablation study on the components of HAS}

\begin{table*}[t!]
  \centering
  \small
  \caption{Ablation study of HAS components on InternVL3.}
  \label{tab:component_ablation}
  \begin{tabular*}{\textwidth}{@{\extracolsep{\fill}} l c c c c c c c c c}
    \toprule 
    \multirow{2}{*}{Model} & \multirow{2}{*}{Frames} 
    & \multicolumn{2}{c}{LVB} 
    & \multicolumn{3}{c}{VideoMME} 
    & \multicolumn{3}{c}{MLVU} \\
    \cmidrule(lr){3-4} \cmidrule(lr){5-7} \cmidrule(lr){8-10} 
    & & 600 & 3600 & Short & Med. & Long & Hol. & S-D & M-D \\
    \midrule
    TOP    & 64 & 63.6 & 56.2 & 77.8 & 63.6 & 54.6 & 80.8 & 75.4 & 57.0 \\
    DPP    & 64 & 63.8 & 55.7 & \underline{78.0} & 64.8 & 55.4 & 82.2 & 76.3 & 56.8 \\
    DPP+GC & 64 & \underline{64.3} & 56.0 & 77.9 & \underline{68.3} & \textbf{57.0} & \textbf{82.3} & 77.1 & 57.9 \\
    DPP+LC & 64 & 61.7 & \textbf{56.4} & \underline{78.0} & 67.0 & 55.2 & \textbf{82.3} & \textbf{77.9} & \textbf{59.1} \\
    HAS    & 64 & \textbf{64.6} & \textbf{56.4} & \textbf{78.1} & \textbf{70.0} & \underline{56.3} & \textbf{82.3} & \underline{77.6} & \underline{58.1} \\
    \bottomrule
  \end{tabular*}
\end{table*}

To validate the contribution of each component in HAS, we conducted a detailed ablation study, with results presented in Table~\ref{tab:component_ablation}. Our ablation study (Table~\ref{tab:component_ablation}) begins by comparing DPP against TOP. DPP improves over TOP on most metrics, with notable gains on benchmarks like VideoMME-Med (64.8 vs. 63.6) and MLVU HoL (82.2 vs. 80.8). This superiority stems from DPP's inherent ability to balance both relevance and diversity. Therefore, we select DPP for our backbone frame sampling.


Building on this, we observe that different context types cater to different benchmark demands. Specifically, adding Global Context (DPP+GC) yields substantial gains on benchmarks emphasizing global semantics, boosting the VideoMME-Med score from 64.8 to 68.3. Conversely, Local Context (DPP+LC) proves highly effective for tasks requiring fine-grained detail, achieving excellent scores on MLVU S-D 77.9 and M-D 59.1. These results show that GC and LC each offer unique, complementary advantages. This directly motivates our adaptive approach.


By successfully synthesizing these complementary strengths, our full method, HAS, achieves the best or second-best performance across all benchmarks. It secures top scores of 64.6 on LVB-600 and 70.0 on VideoMME-Med, and remains highly competitive even where not strictly the top performer, such as on VideoMME-Long (56.3). In a few cases, a single-context variant (e.g., DPP+LC or DPP+GC) can be slightly better, likely because some benchmarks strongly prefer either local or global context. Overall, HAS provides a strong balance between high performance and generalization, avoiding large swings across benchmarks.

\subsubsection{Ablation study of Alternative Strategies for Global Context Sampling.}
\label{ablation_study_GC}

In HAS, a portion of the budget is dedicated to Global Context to provide a global perspective of the video. To identify the most effective GC sampling strategy, we evaluated four approaches: (\romannumeral 1)Uniform (UNI), which samples frames at uniform temporal intervals; (\romannumeral 2)Max-in-Segment (MiS), which divides the video into uniform segments and selects the most relevant frame to the query from each; (\romannumeral 3)Clustering-based (CLUS), which performs clustering within each uniform segment and then selects the centroid of the largest cluster; and (\romannumeral 4)Temporal Bridging (TB), which iteratively samples the midpoint of the largest temporal gap. 
As shown in Table~\ref{tab:ablation_gc}, the simple UNI proves highly effective, with leading performance in benchmarks such as LVB and VideoMME.
We attribute this to a mismatch: Global Context aims to offer a broad, unbiased temporal overview, whereas methods like MiS and CLUS introduce relevance or centrality biases. These biases conflict with this goal and compromise the uniform coverage vital for a holistic view. Therefore, we conclude that a simple unbiased strategy like Uniform sampling is the most robust choice for supplementing a query-focused backbone.

\subsubsection{Ablation study of different fps }

\begin{table*}[t]
  \centering
  \small
  \setlength{\tabcolsep}{6pt}
  \renewcommand{\arraystretch}{0.95}

  \begin{minipage}[t]{0.48\textwidth}
    \centering
    \caption{Ablation study of GC strategies.}
    \label{tab:ablation_gc}
    \vspace{0.3em}
    \begin{tabular*}{\linewidth}{@{\extracolsep{\fill}}lccc@{}}
      \toprule
      Method & LVB & VideoMME & MLVU \\
      \midrule
      UNI  & \textbf{62.2} & \textbf{68.1} & 75.8 \\
      MiS  & 61.0          & 67.4          & \textbf{76.0} \\
      CLUS & 61.1          & 67.9          & 75.7 \\
      TB   & 61.9          & 67.4          & 75.2 \\
      \bottomrule
    \end{tabular*}
  \end{minipage}\hfill%
  \begin{minipage}[t]{0.48\textwidth}
    \centering
    \caption{Ablation study of different FPS.}
    \label{tab:perf_vs_fps}
    \vspace{0.3em}
    \begin{tabular*}{\linewidth}{@{\extracolsep{\fill}}lccc@{}}
      \toprule
      FPS & LVB & VideoMME & MLVU \\
      \midrule
      2    & \textbf{62.3} & \textbf{68.7} & \textbf{76.1} \\
      1    & 62.2          & 68.1          & 75.8 \\
      0.5  & 61.3          & 68.0          & 75.3 \\
      0.25 & 61.1          & 67.6          & 75.4 \\
      \bottomrule
    \end{tabular*}
  \end{minipage}
\end{table*}

Table~\ref{tab:perf_vs_fps} reports the performance of HAS under different pre-sampling FPS. Using a higher FPS generally yields better accuracy, as a denser candidate pool provides more fine-grained temporal coverage and is less likely to miss critical moments or key evidence. In our main experiments, we set the pre-sampling rate to 1~FPS to balance performance with efficiency, since computing query-frame relevance scores becomes increasingly expensive as FPS grows.

\subsubsection{Ablation study of hyperparameters}
\label{ablation_study_hyperpara}

\begin{figure}[h]
    \centering
    \begin{subfigure}[b]{0.23\textwidth}
        \centering
        \includegraphics[width=\linewidth]{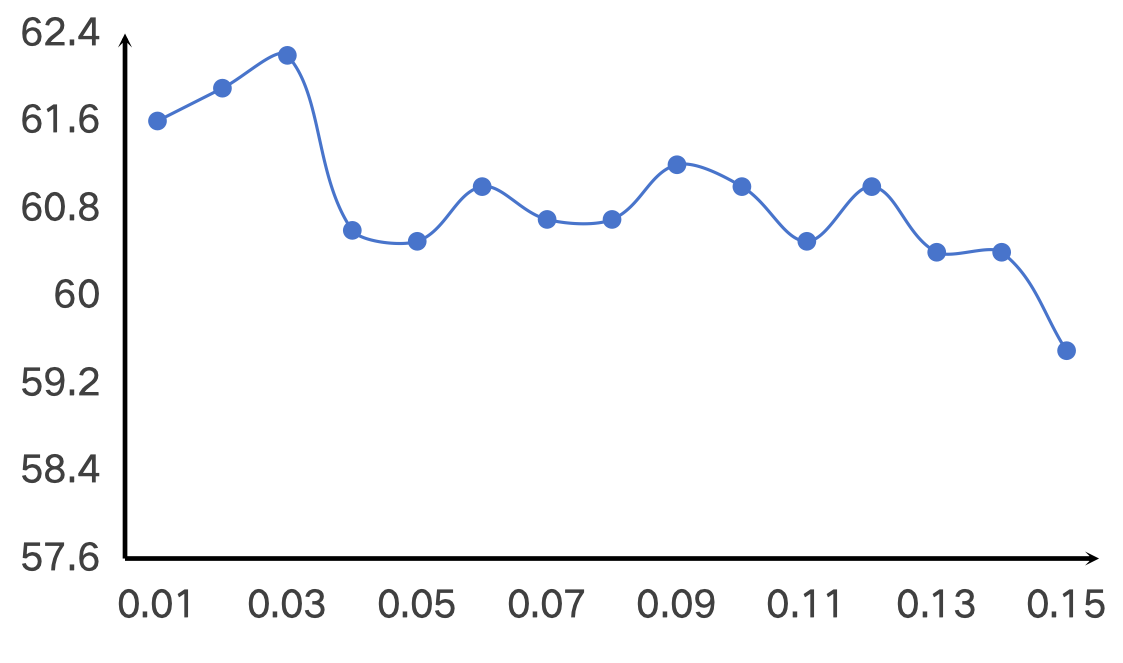} 
        \caption{Perf. on LVB}
        \label{fig:sub_a_final}
    \end{subfigure}
    \hspace{0.01\textwidth}%
    \begin{subfigure}[b]{0.23\textwidth}
        \centering
        \includegraphics[width=\linewidth]{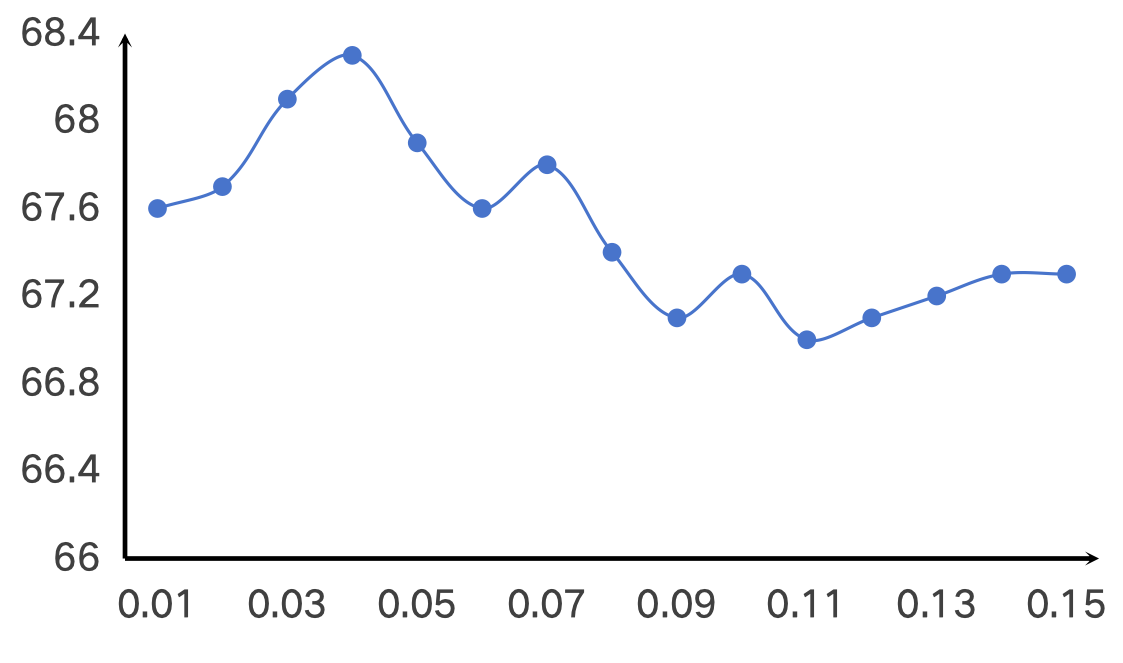} 
        \caption{Perf. on VMME}
        \label{fig:sub_b_final}
    \end{subfigure}
    \hspace{0.01\textwidth}%
    \begin{subfigure}[b]{0.23\textwidth}
        \centering
        \includegraphics[width=\linewidth]{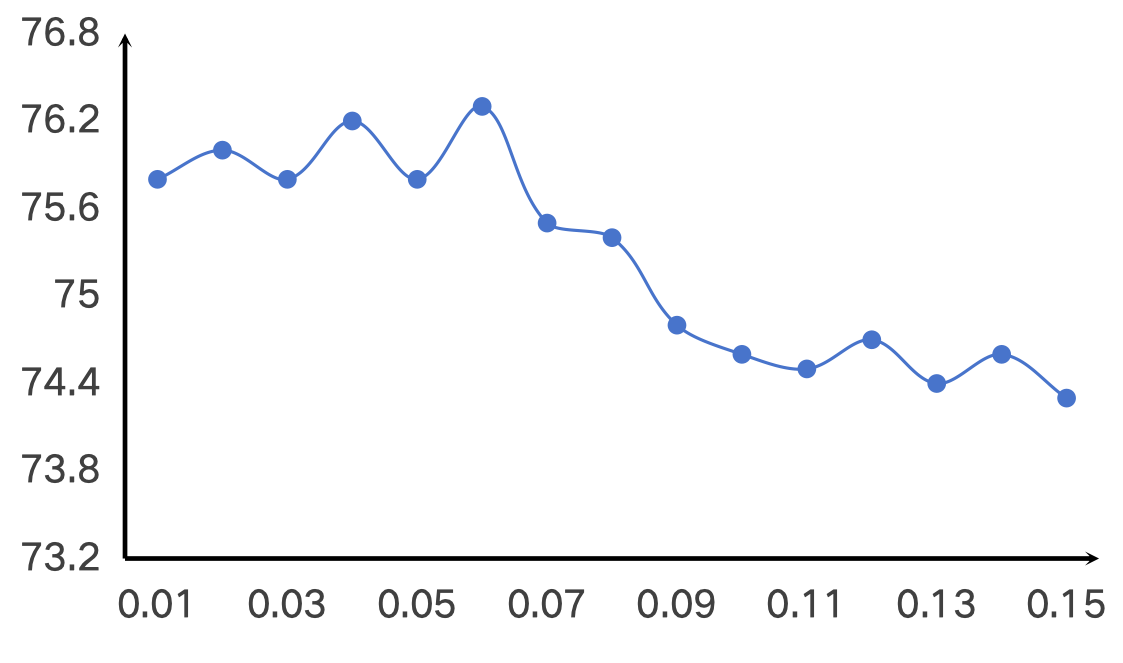} 
        \caption{Perf. on MLVU}
        \label{fig:sub_c_final}
    \end{subfigure}
    \hspace{0.01\textwidth}%
    \begin{subfigure}[b]{0.23\textwidth}
        \centering
        \includegraphics[width=\linewidth]{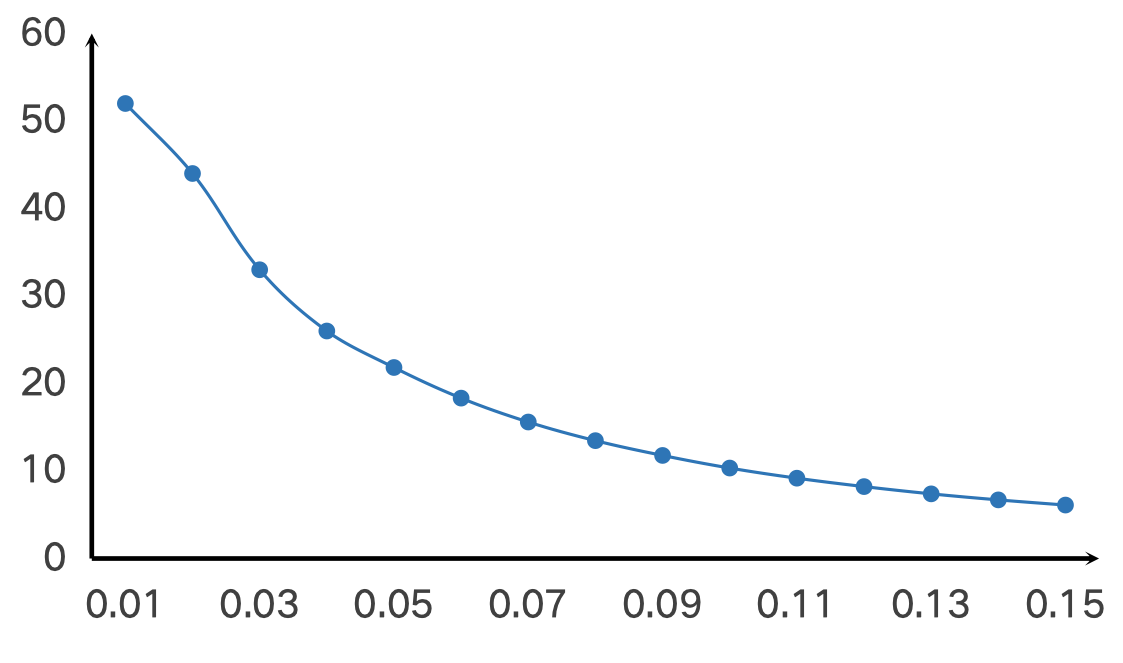} 
        \caption{$|\mathcal{I}_{\text{backbone}}|$ on VMME}
        \label{fig:sub_d_final}
    \end{subfigure}

    \caption{
Analysis of the hyperparameter $\eta$. (a)--(c) show the model performance under different values of $\eta$ on three benchmarks, and (d) shows the number of selected backbone frames under different values of $\eta$. 
    }
    \label{fig:hyperparameters}
\end{figure}


In this section, we analyze $\eta$, the marginal-gain threshold in our DPP-based backbone selection (smaller $\eta$ selects more frames). As shown in Figure~\ref{fig:hyperparameters}, the performance exhibits a consistent trend across benchmarks, first increasing and then decreasing. 
With a low $\eta$, the backbone may include low-relevance frames and leave too little budget for the second-stage enrichment. With a high $\eta$, too few backbone frames are selected, causing the model to miss some key moments. As a result, performance is best at a moderate $\eta$ (around $\eta=0.03$--$0.04$), which strikes a good balance between capturing sufficient key moments and reserving enough budget for contextual enrichment. This behavior further supports the necessity of our two-stage design. We therefore set $\eta=0.03$ in all main experiments. The sensitivity to the similarity threshold $\tau$ is further analyzed in Appendix~\ref{tab:tau_sensitivity}.

\vspace{-1.5ex}
\section{Conclusion}

In this work, we find that existing video frame sampling methods are often governed by a one-sided principle, making it hard to simultaneously meet the heterogeneous evidential needs of different queries. To address this limitation, we propose HAS, a hierarchical framework that dynamically tailors sampled frames to the specific needs of each query. HAS first constructs a query-aware backbone via DPP to capture diverse and relevant key moments, and then adaptively allocates the remaining sampling budget between Global Context (to cover the broader video timeline for holistic queries) and Local Context (to capture fine-grained temporal dynamics for moment-centric queries). With this adaptive design, HAS consistently delivers strong performance across multiple benchmarks and a range of MLLMs.


\bibliographystyle{plainnat}
\bibliography{main}




\end{document}